\documentclass[letterpaper, 10 pt, journey]{IEEEtran}
\usepackage{amsmath,amsfonts}
\usepackage{subfigure}
\usepackage{algorithmic}
\usepackage{algorithm}
\usepackage{array}
\usepackage[caption=false,font=normalsize,labelfont=sf,textfont=sf]{subfig}
\usepackage{textcomp}
\usepackage{stfloats}
\usepackage{url}
\usepackage{multirow}
\usepackage{verbatim}
\usepackage{graphicx}
\usepackage{gensymb}
\usepackage{soul}
\usepackage{booktabs}
\usepackage{cite}
\begin{document}

\title{LiDAR-Inertial 3D SLAM with Plane Constraint for Multi-story Building}

\author{Jiashi Zhang, Chengyang Zhang, Jun Wu, Jianxiang Jin, Qiuguo Zhu
\thanks{*This work was supported by NSFC 62088101 Autonomous Intelligent Unmanned Systems and the National Key R\&D Program of China (2020YFB1313300) and National Quality Infrastructure Research Program of Zhejiang Administration for Market Regulation (20190104).}
\thanks{Jiashi Zhang, Chengyang Zhang, Jun Wu, Jianxiang Jin, Qiuguo Zhu are with the State Key Laboratory of Industrial Control and Technology, Zhejiang University, Institute of Cyber-System and Control, Zhejiang University, Hangzhou, P.R. China. Qiuguo Zhu is the corresponding author (e-mail: qgzhu@zju.edu.cn).}
}




\maketitle

\begin{abstract}
The ubiquitous planes and structural consistency are the most apparent features of indoor multi-story Buildings compared with outdoor environments. In this paper, we propose a tightly coupled LiDAR-Inertial 3D SLAM framework with plane features for the multi-story building. The framework we proposed is mainly composed of three parts: tightly coupled LiDAR-Inertial odometry, extraction of representative planes of the structure, and factor graph optimization. By building a local map and inertial measurement unit (IMU) pre-integration, we get LiDAR scan-to-local-map matching and IMU measurements, respectively. Minimize the joint cost function to obtain the LiDAR-Inertial odometry information. Once a new keyframe is added to the graph, all the planes of this keyframe that can represent structural features are extracted to find the constraint between different poses and stories. A keyframe-based factor graph is conducted with the constraint of planes, and LiDAR-Inertial odometry for keyframe poses refinement. The experimental results show that our algorithm has outstanding performance in accuracy compared with the state-of-the-art algorithms.
\end{abstract}


\section{INTRODUCTION}


With the development of quadruped robots' motion control and environmental perception capabilities, the scenarios they can explore are also expanding from 2D to 3D compared with wheeled mobile robot. Accurate state estimation and mapping are the basic premises for applying robots in the real world. As for indoor environments, especially multi-story buildings, the multi-planar feature can help the robot achieve low-drift simultaneous localization and mapping (SLAM) due to the structural consistency.

\begin{figure}[!t]
\centering
\includegraphics[width=3.4in]{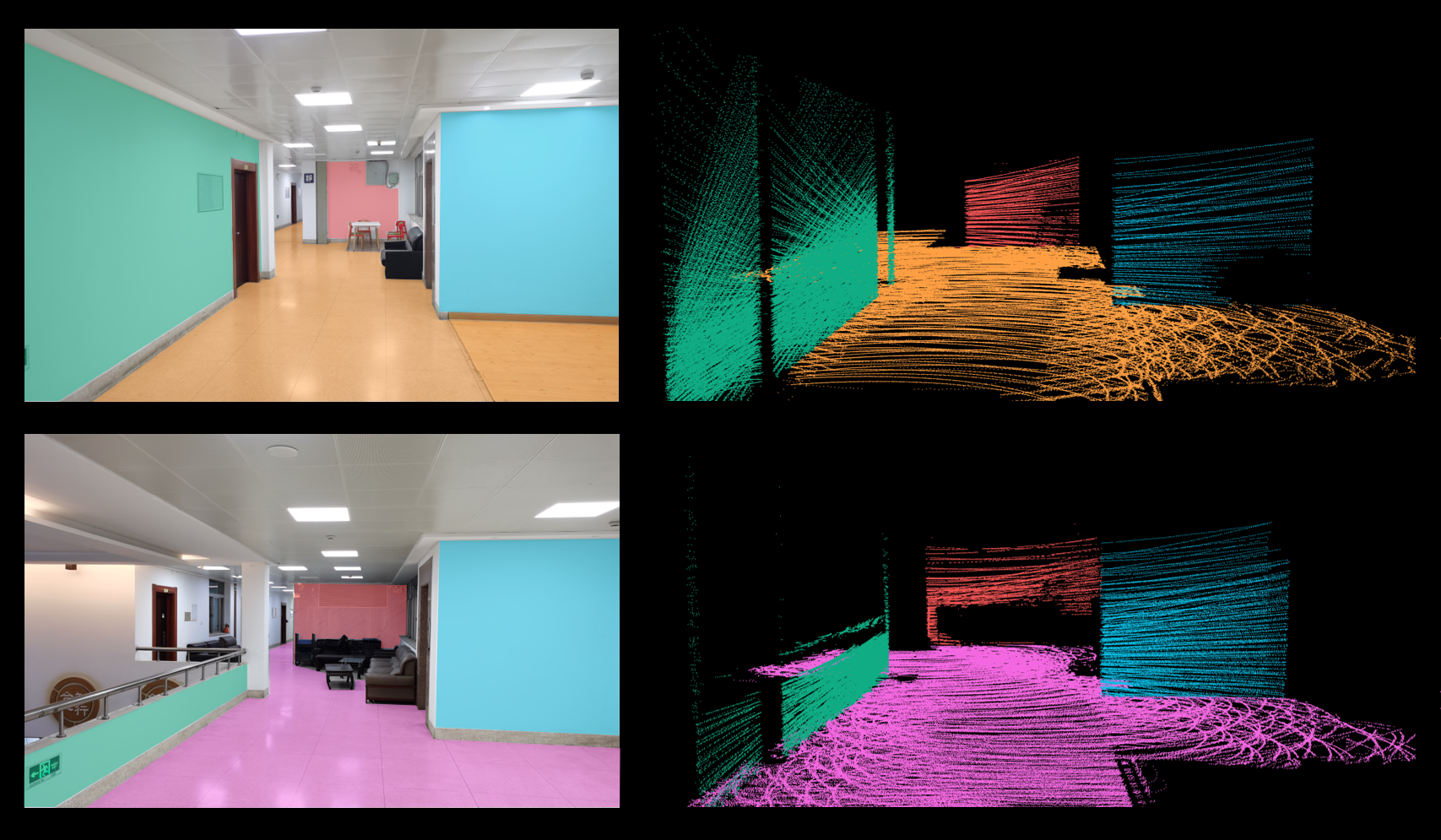}
\caption{Schematic diagram of SRP. On the left are actual scenes from different stories. Different colors represent different SRPs, and the same color represents the same plane. On the right is the SRP extracted from the LiDAR point cloud. Using the same SRP to construct constraints on different stories can eliminate accumulated errors.}
\label{schematic-diagram}
\end{figure}

A 3D LiDAR based on scanning mechanism has the advantages of textureless, invariant to the illumination, and broad horizontal of view (FOV) of 360{\degree}, which is generally used in indoor environments\cite{hesch2010laser, zou2021comparative}. Under normal circumstances, LiDAR-aided SLAM mainly uses extracting corner points and surf points method \cite{zhang2014loam, shan2018lego, shan2020lio, lin2020loam}, Normal Distributions Transform (NDT)\cite{magnusson2007scan} scan matching, or floor extraction\cite{koide2018portable} methods to achieve SLAM for a single floor. Although many algorithms implement SLAM by extracting planes in indoor environments, most only use plane constraints in the odometry part and achieve accurate SLAM algorithms by finding the scan-to-scan plane correspondence. However, when the robot explores from bottom to top in a multi-story building, the existing algorithms cannot achieve accurate state estimation on the robot's 6-DOF, due to long-distance and loop closure does not work. In the multi-story SLAM, due to the consistency of structure between different floors, some planes on different floors can represent the same building structure. Here we call these planes structural representative planes (SRP). Fig. \ref{schematic-diagram} shows an example of SRP, where the same SRP is displayed in the same color on different stories. Finding the correspondence between SRP within the scope is the key to achieving low-drift SLAM in multi-story scenes.

This paper presents a tightly-coupled LiDAR-Inertial 3D SLAM framework using planes to build global constraints. Our framework has three parts: tightly coupled LiDAR-Inertial odometry, extraction of representative planes of the structure, and factor graph optimization. The odometry is obtained by jointly optimizing the relative pose of the scan-to-local-map and the inertial measurement unit (IMU) pre-integration measurements. According to the odometry information, all the SRP will be extracted as candidates for the global plane constraint once a new keyframe is selected. Transform the global SRP to the keyframe coordinate system, and construct the global constraint relationship between keyframes according to the direction of planes' normals and the distance to the coordinate origin. Add odometry information and constraint information from planes to the factor graph, perform global optimization, and get the accurate pose of each keyframe. The main contributions of this paper can be summarized as follows:

\begin{itemize}
\item{We propose the method of finding and constructing the global constraints of SRP in the multi-story blocks to achieve accurate 6-DOF state estimation of the robot when the loop closure is not possible.}
\item{We propose a tightly coupled LiDAR-Inertial, keyframe-based SLAM framework to get the dense 3D point cloud maps of multi-story blocks.}
\item{We validate the algorithm using the data collected from Velodyne VLP-16 and Xsens Mti-300 mounted on a real quadruped robot (Jueying Robot). Compared with other state-of-the-art algorithms, better results are obtained.}
\end{itemize}

\section{RELATED WORK}

\textbf{LiDAR Inertial odometry} In recent years, 3D LiDAR and IMU have been widely used in SLAM, both indoors and outdoors. The fusion methods of LiDAR and IMU are mainly divided into two categories: loosely coupled and tightly coupled. In the field of loosely coupled, LOAM \cite{zhang2014loam} is a classic loosely coupled framework. It uses the orientation calculated by the IMU de-skew the point cloud and as prior information in the optimization process. The same method is also applied to its variants LeGO-LOAM \cite{shan2018lego}. Zhen W \emph{et al.} \cite{zhen2017robust} integrate IMU measurements and LiDAR estimations from a Gaussian particle filter (GPF) and a pre-built map with error state Kalman filter (ESKF). The more popular loosely coupled method is the extended Kalman filters (EKF). \cite{lynen2013robust, gao2015ins, demir2019robust} propose some generic EKF-based frameworks for robot state estimation, which can integrate the measurements of LiDAR and IMU, as well as global position system (GPS). LIO-Mapping \cite{ye2019tightly} realized LiDAR-Inertial tightly coupled algorithm by optimizing the cost function that includes both LiDAR and inertial measurements. However, the optimization process is carried out in a sliding window, so the time-consuming calculations make it impossible to maintain real-time performance. In their follow-up work, R-LINS \cite{qin2020lins}, they use iterated-ESKF for the first time to achieve LiDAR-Inertial tightly coupled fusion and propose an iterated Kalman filter \cite{bell1993iterated} to reduce wrong matchings in each iteration. A tightly coupled framework based on iterated Kalman filter is presented in \cite{xu2021fast}, similar to R-LINS. An incremental k-d tree data structure is adopted to ensure cumulative updates and dynamic balance to ensure fast and robust LiDAR mapping. LIO-SAM \cite{shan2020lio} proposed by Shan T \emph{et al.} optimizes the measurements of LiDAR and IMU by factor graph, and at the same time, estimates the bias of the IMU.

\textbf{SLAM related to plane features} Whether in vision-based SLAM or LiDAR-based SLAM, plane-related features are widely used to improve state estimation accuracy. In LiDAR-based SLAM, LOAM \cite{zhang2014loam} proposed extracting feature points from planar surface patches and sharp edges based on curvature calculation and improved the iterative closest point (ICP) \cite{rusinkiewicz2001efficient} method based on the extracted feature points demonstrating the superb LiDAR odometry effect. Koide K \emph{et al.} \cite{koide2018portable} realize SLAM in a large-scale environment by detecting the ground, assuming that the indoor environment is a single flat floor. But this assumption is not applicable in all scenes and can only limit the height on the z-axis. Kaess \cite{kaess2015simultaneous} introduces an efficient parametrization of planes based on quaternion, suitable for least-squares estimation. Besides, he presents a relative plane formulation to speed up the convergence process. LIPS \cite{geneva2018lips} extract the plane in the three-axis direction of the point cloud, not only the ground plane, and combine the plane and IMU measurements in a graph-based framework. At the same time, the closets point (CP) is used to represent the plane to solve the singularity. K Pathak \emph{et al.} \cite{pathak2010fast} present a new algorithm called minimally uncertain maximal consensus (MUMC) to determine the unknown plane correspondences in the front-end. \begin{math}\pi\end{math}-LSAM, an indoor environment SLAM system using planes as landmarks, was proposed by Zhou L \emph{et al.} \cite{zhou2021pi}. They adopt plane adjustment (PA) as the back-end to optimize plane parameters and poses of keyframes, similar to bundle adjustment (BA) in visual SLAM. Their subsequent work \cite{zhou2021lidar} extended this by using first-order Taylor expansion to replace the Levenberg Marquardt (LM) \cite{more1978levenberg} method. To achieve faster computational speed, they define the integrated cost matrix (ICM) for each plane and achieve outstanding SLAM effects in a single-layer indoor environment. All of the above frameworks use a single LiDAR or a loosely coupled method of LiDAR and IMU as the front-end. On the contrary, we use a tightly coupled LiDAR-Inertial method as the front-end, which can obtain a more accurate prior pose of the keyframe, making it more precise when looking for the corresponding between the planes.

\section{SYSTEM OVERVIEW}
\begin{figure*}[!t]
\centering
\includegraphics[width=6in]{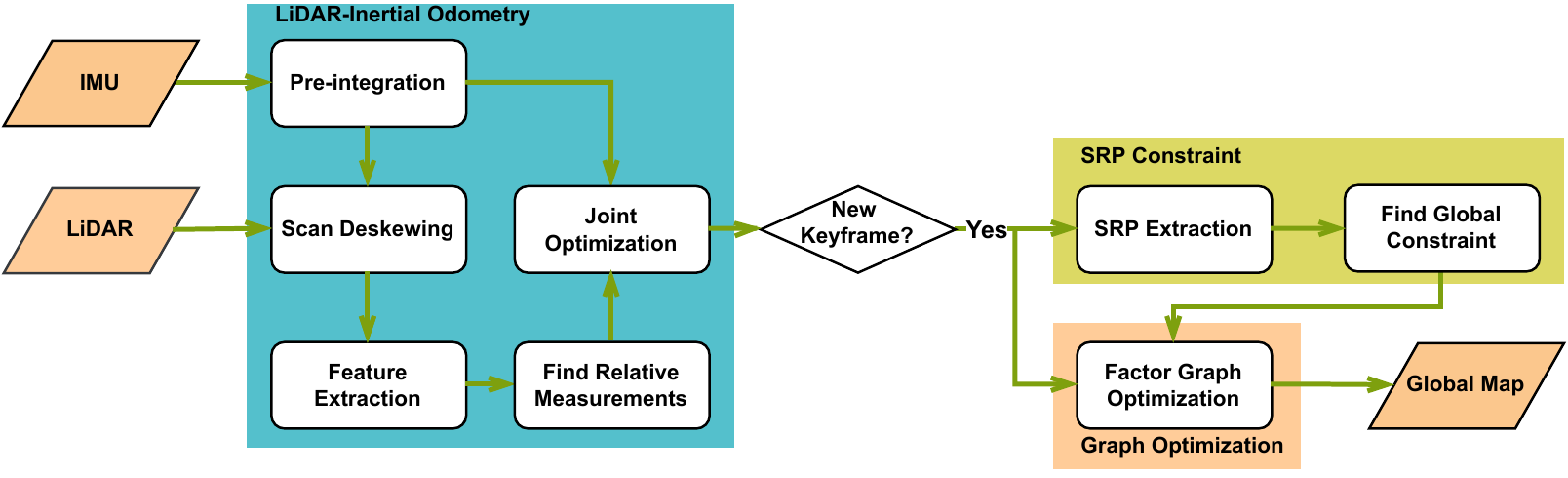}
\caption{System overview of our algorithm.}
\label{system_overview}
\end{figure*}

Our algorithm consists of three parts, LiDAR-Inertial Odometry, SRP Constraint and Graph Optimization, as shown in Fig. \ref{system_overview}.

The LiDAR-Inertial Odometry performs pre-integration on high-frequency IMU data and corrects the motion distortion of the point cloud. Find the relative measurements of LiDAR by extracting feature points from the point cloud. Optimize the cost function that includes pre-integration and LiDAR relative measurements to obtain the odometry.

The SRP Constraint extracts SRP from the point cloud of each keyframe and matches with the global SRP to form the edge between the vertices in the factor graph.

The graph optimization performs optimization every time a new SRP constraint is formed and adjusts the poses of the keyframes.




\section{LIDAR-INERTIAL ODOMETRY}
\subsection{IMU Pre-integration}
The LiDAR and IMU reference frames at time $t$ are noted 
${L_t}$ and ${I_t}$, respectively. The state $\mathbf{X}^{W}_{I_t}$ of IMU to be estimated in the world frame ${W}$ and the extrinsic matrix $\mathbf{T}_I^L$ from IMU to LiDAR can be written as:
\begin{align}
\begin{split}
\mathbf{X}_{I_{t}}^{W} & =\left[\begin{array}{lllll}
{\mathbf{p}_{I_{t}}^{W}}^{T} & {\mathbf{v}_{I_{t}}^{W}}^{T} & {\mathbf{q}_{I_{t}}^{W}}^{T} & {\mathbf{b}_{a_{t}}}^{T} & {\mathbf{b}_{g_{t}}}^{T}
\end{array}\right]^{T}
\\
\mathbf{T}_{I}^{L} & =\left[\begin{array}{ll}
{\mathbf{p}_{I}^{L}}^{T} & {\mathbf{q}_{I}^{L}}^{T}
\end{array}\right]^{T}
\end{split}
\end{align}
where $\mathbf{p}_{I_{t}}^{W}$, $\mathbf{v}_{I_{t}}^{W}$, and $\mathbf{q}_{I_{t}}^{W}$ are the position, velocity, and orientation of IMU in the world frame ${W}$ at time $t$. $\mathbf{b}_{a_{t}}$ and $\mathbf{b}_{g_{t}}$ are the bias of accelerometer and gyroscope of IMU.

Using $\hat{\mathbf{a}}_t$ and $\hat{\boldsymbol{\omega}}_t$ to represent the raw measurements of acceleration and angular velocity in frame ${L_t}$, respectively, and they can be defined as:
\begin{align}\label{raw-IMU-measurement}
\begin{split}
\hat{\mathbf{a}}_{t}&=\mathbf{R}_{W}^{I_{t}}\left(\mathbf{a}_{i}^{W}-\mathbf{g}^{W}\right)+\mathbf{b}_{a_{t}}+\mathbf{n}_{a}
\\
\hat{\boldsymbol{\omega}}_{t}&=\boldsymbol{\omega}_{t}+\mathbf{b}_{w_{t}}+\mathbf{n}_{w}
\end{split}
\end{align}
where $\mathbf{R}_{W}^{I_{t}}$ is the rotation matrix from world frame ${W}$ to frame ${I_t}$ and $\mathbf{g}^{W}$ is the constant gravity vector in world frame ${W}$. $\mathbf{n}_{a}$ and $\mathbf{n}_{w}$ are the noises of IMU, modeled as Gaussian white noises.

With the continuous-discrete IMU inputs, the position, velocity, and orientation of the IMU at time $t+\Delta t$ can be calculated according to Eq. (\ref{raw-IMU-measurement}).
\begin{align}
    \begin{split}
        \mathbf{p}_{t+\Delta t}&=\mathbf{p}_{t}+\mathbf{v}_{t} \Delta t+\frac{1}{2}\mathbf{g}^W{\Delta t}^2
        +\frac{1}{2}\mathbf{R}_{I_t}^W\left(\hat{\mathbf{a}}_{t}-\mathbf{b}_{a_{t}}-\mathbf{n}_{a}\right){\Delta t}^2
        \\
        \mathbf{v}_{t+\Delta t}&=\mathbf{v}_{t}+\mathbf{g}^W{\Delta t}+\mathbf{R}_{I_t}^W\left(\hat{\mathbf{a}}_{t}-\mathbf{b}_{a_{t}}-\mathbf{n}_{a}\right){\Delta t}
        \\
        \mathbf{q}_{t+\Delta t}&=\mathbf{q}_{t} \otimes  \frac{1}{2} \Omega\left(\hat{\boldsymbol{\omega}}_{t}-\mathbf{b}_{w_{t}}-\mathbf{n}_{w}\right) \mathbf{q}_{t} \Delta t
        \\
    \end{split}
\end{align}
where $\otimes$ is used for the multiplication of two quaternions, and $\Omega(\boldsymbol{\omega})$ is:
\begin{align}
    \begin{split}
        \Omega(\boldsymbol{\omega}) = \left[\begin{array}{cc}
-\lfloor\boldsymbol{\omega}\rfloor_\times & \boldsymbol{\omega} \\
-\boldsymbol{\omega}^{T} & 0
\end{array}\right]
    \end{split}
\end{align}
where $\lfloor\cdot\rfloor_\times \in \mathbb{R}^{3\times 3}$ stands for the skew-symmetric matrix, let $t_i$ and $t_j$ be the starting time and ending time of a raw LiDAR scan $\Tilde{\mathcal{S}_i}$, respectively, so the pre-integration measurements $\Delta \mathbf{p}_{ij}$, $\Delta \mathbf{v}_{ij}$, $\Delta \mathbf{q}_{ij}$ of IMU from time $t_i$ to $t_j$ can be computed as:
\begin{align}\label{IMU-measurement}
    \begin{split}
        \Delta \mathbf{p}_{i j}&=\sum_{k=i}^{j-1}\left[\Delta \mathbf{v}_{i k} \Delta t+\frac{1}{2} \Delta \mathbf{R}_{i k}\left(\hat{\mathbf{a}}_{k}-\mathbf{b}_{a_{k}}-\mathbf{n}_{a}\right) \Delta t^{2}\right]
        \\
        \Delta \mathbf{v}_{i j}&=\sum_{k=i}^{j-1} \Delta \mathbf{R}_{i k}\left(\hat{\mathbf{a}}_{k}-\mathbf{b}_{a_{k}}-\mathbf{n}_{a}\right) \Delta t
        \\
        \Delta \mathbf{q}_{i j}&=\prod_{k=i}^{j-1} \delta \mathbf{q}_{k}=\prod_{k=i}^{j-1}\left[\begin{array}{c}
\frac{1}{2} \Delta t\left(\hat{\boldsymbol{\omega}}_{k}-\mathbf{b}_{w_{k}}-\mathbf{n}_{w}\right) \\
1
\end{array}\right]
    \end{split}
\end{align}
Readers can refer to \cite{forster2016manifold} for detailed derivation from Eq. (\ref{raw-IMU-measurement}) to Eq. (\ref{IMU-measurement}).
\subsection{Scan Deskewing and Feature Extraction}

Due to the relative movement between the laser and the robot, there will be motion distortion for the raw LiDAR output $\Tilde{\mathcal{S}}_i$, where $\Tilde{\mathcal{S}}_i$ represents the point cloud starting from time $t_i$ to time $t_j$. Every point $\mathbf{x}(t) \in \Tilde{\mathcal{S}}_i$ is transformed to the correct position by linear interpolation to $\mathbf{T}_{ij}^L$ according to its timestamp, where $t\in[t_i, t_j)$. $\mathbf{T}_{ij}^L$ can be obtained by IMU pre-integration and extrinsic matrix $\mathbf{T}_I^L$, and the undistorted scan can be represented by $\mathcal{S}_i$.

To improve the efficiency of calculation, only the feature points that can reflect the characteristics of the surrounding environment are selected to find the relative pose of the LiDAR. Here we use the method of extracting feature points located on sharp edges and planar surfaces proposed by LOAM. The extracted edge and planar feature points from $\mathcal{S}_i$ are denoted as $\mathcal{F}^{L_i}_e$ and $\mathcal{F}^{L_i}_p$, respectively.

\subsection{LiDAR Relative Measurements}
When the new feature points $\mathcal{F}^{L_i}_e$ and $\mathcal{F}^{L_i}_p$ are extracted, the measurements of LiDAR need to be found to jointly perform the optimization with IMU.
\subsubsection{Building Local Map}
Since the points of a single scan are not dense enough, to obtain more accurate LiDAR measurements, we use a sliding window to construct a local map. The sliding window contains $n$ LiDAR frames from time $t_{i-1}$ to time $t_{i-n}$. Since we have extracted planar points and edge points separately, we transform $\left\{\mathcal{F}^{L_{i-n}}_e, ..., \mathcal{F}^{L_{i-2}}_e, \mathcal{F}^{L_{i-1}}_e\right\}$ and $\left\{\mathcal{F}^{L_{i-n}}_p, ..., \mathcal{F}^{L_{i-2}}_p, \mathcal{F}^{L_{i-1}}_p\right\}$ to frame ${L_{i-1}}$ respectively with $\left\{\mathbf{T}_{i-n}^{i-1}, ..., \mathbf{T}_{i-2}^{i-1}, \mathbf{T}_{i-1}^{i-1}\right\}$ to obtain two feature local maps, $\mathcal{M}_{e}^{L_{i-1}}$ and $\mathcal{M}_{p}^{L_{i-1}}$. They will be downsampled to get the centroid points in each 3D voxel grid to remove duplicate points.
\subsubsection{Scan Matching}
The relationship between the feature points and the local maps at time $t_i$ can be calculated by the point-line and the point-plane distances. First, transform the feature points $\mathcal{F}^{L_i}_e$ and $\mathcal{F}^{L_i}_p$ to frame ${L}_{i-1}$. The prediction transformation $\mathbf{T}_{i-1}^{i}$ used here is obtained through IMU pre-integration and extrinsic matrix $\mathbf{T}_I^L$. Here we take the plane points as an example. For each transformed plane point $^{\prime}\mathbf{x}_p^{L_i}$, find the nearest $m$ points in $\mathcal{M}_{p}^{L_{i-1}}$ to fit a plane in the frame ${L}_{i-1}$ and express in Hesse normal form:
\begin{align}
\label{hesse}
    \begin{split}
        \mathbf{x}^{T}\mathbf{n}-d=0
    \end{split}
\end{align}
where $\mathbf{n}$ is the unit normal vector of plane, and $d$ is the distance from plane to the origin of frame ${L}_{i-1}$. So for each plane point $\mathbf{x}_p^{L_i}\in\mathcal{F}_p^{L_i}$, the residual can be expressed as the point-plane distance:
\begin{align}
    \begin{split}
        \mathbf{T}_{L_i}^{L_{i-1}} = & \begin{bmatrix} \mathbf{R}_{L_i}^{L_{i-1}} & \mathbf{p}_{L_i}^{L_{i-1}} \\ 
            \mathbf{0} &  1 \end{bmatrix}\\
        {r}_\mathcal{P}(\mathbf{T}_{L_i}^{L_{i-1}}) =& \left(\mathbf{R}_{L_i}^{L_{i-1}}\mathbf{x}_{L_i}^p + \mathbf{p}_{L_i}^{L_{i-1}}\right)^{T}\mathbf{n}-d
    \end{split}
\end{align}
Similar to the calculation method of the plane point, the Hesse normal form can also describe the line in $\mathbb{R}^2$. For each edge point, the residual can be represented as the point-line distance:
\begin{align}
    \begin{split}
        {r}_\mathcal{E}(\mathbf{T}_{L_i}^{L_{i-1}}) = \left(\mathbf{R}_{L_i}^{L_{i-1}}\mathbf{x}_{L_i}^e + \mathbf{p}_{L_i}^{L_{i-1}}\right)^{T}\mathbf{n}-d
    \end{split}
\end{align}
\subsection{Front-End Optimization}
We build a cost function including IMU measurements and LiDAR measurements jointly. To get more accurate odometry for each frame of LiDAR, we optimize all the states in the sliding window iteratively. For a sliding window of size $n$ at time $t_i$, the states need to be optimized is $\mathbf{X}_i = \left[\mathbf{T}_{i}^{i-n}, ..., \mathbf{T}_{i-(n-1)}^{i-n}\right]$, and the final cost function is described as:
\begin{align}
    \begin{split}
\underset{\mathbf{X}_i}{min}\frac{1}{2}&\Bigg\{
\sum_{\alpha\in\left\{i-n, ..., i-1\right\}}\left\|{r}_\mathcal{I}({z}_{\alpha+1}^{\alpha}, \mathbf{X}_i)\right\|^{2}_{\mathbf{C}_{I_{\alpha+1}}^{I_{\alpha}}}\\
&+\sum_{\mathbf{x}_{L_i}^{p}\in\mathcal{F}_{L_i}^{p}\atop\beta\in\left\{i-n,...,i-1\right\}}\left\|{r}_{\mathcal{P}}(\mathbf{X}_i)\right\|^{2}_{^{p}\mathbf{C}_{L_{\beta+1}}^{L_{i-n}}}\\
&+\sum_{\mathbf{x}_{L_i}^{e}\in\mathcal{F}_{L_i}^{e}\atop\gamma\in\left\{i-n,...,i-1\right\}}\left\|{r}_\mathcal{E}(\mathbf{X}_i)\right\|^{2}_{^{e}\mathbf{C}_{L_{\gamma+1}}^{L_{i-n}}}
\Bigg\}
    \end{split}
\end{align}
where $\left\|\mathbf{X}\right\|^{2}_{\mathbf{C}} = \mathbf{X}^T\mathbf{C}\mathbf{X}$ and ${r}_\mathcal{I}(\mathbf{X}_i)$ is the residual of IMU measurements, which is defined in\cite{ye2019tightly}. ${r}_{\mathcal{P}}(\mathbf{X}_i)$ and ${r}_{\mathcal{E}}(\mathbf{X}_i)$ are the residuals of planar points matching and edge points matching. $\mathbf{C}_{I_{\alpha+1}}^{I_{\alpha}},\mathbf{C}_{L_{\beta+1}}^{L_{i-n}},\mathbf{C}_{L_{\gamma+1}}^{L_{i-n}}$ represent the covariance matrix. This non-linear least squares problem can be solved using the Levenberg–Marquardt algorithm\cite{more1978levenberg}.

\begin{figure}[!t]
\centering
\includegraphics[width=3.4in]{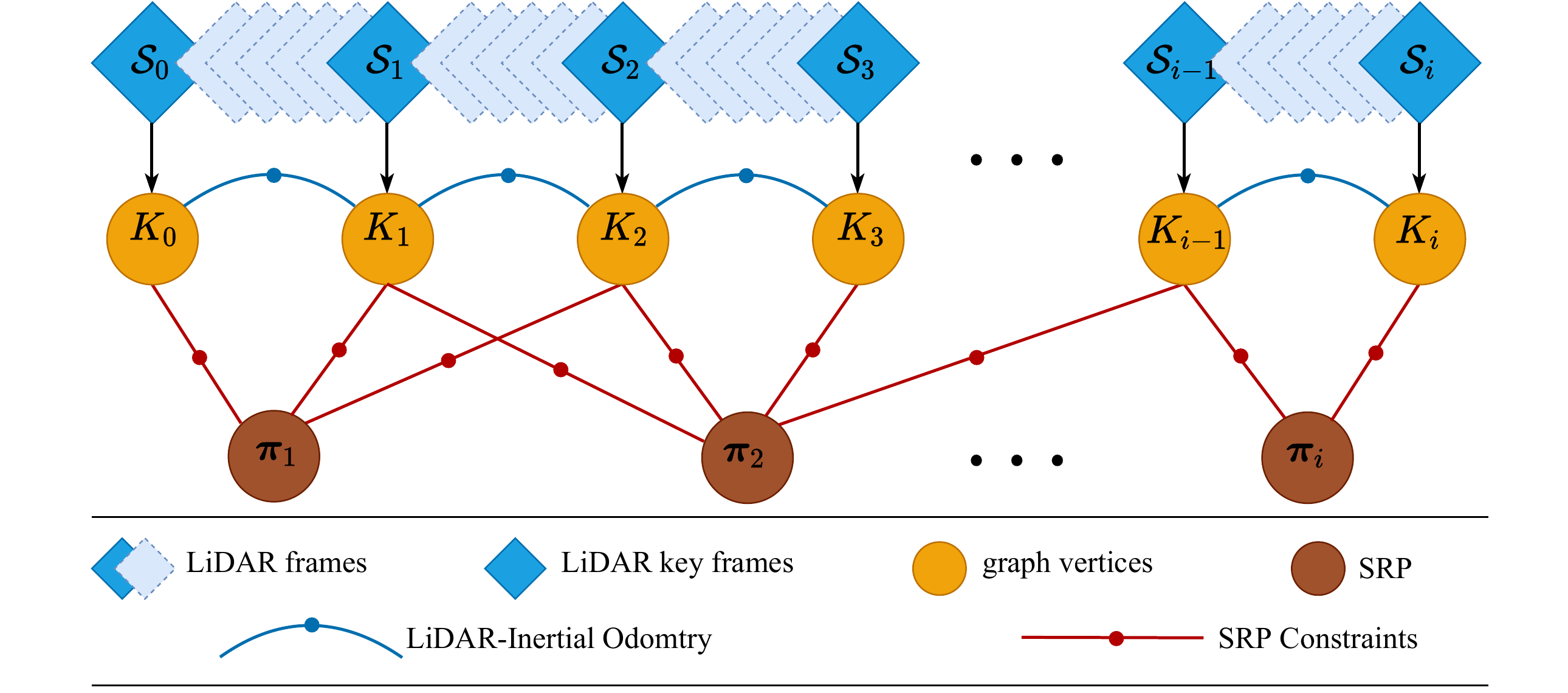}
\caption{The structure of the factor graph. The system selects keyframes based on the odometry as the vertices of the factor graph. The edges between the vertices are formed by LiDAR-Inertial odometry (blue curve) and SRP constraints (red line).}
\label{factor_graph}
\end{figure}

\section{SRP CONSTRAINT AND GRAPH OPTIMIZATION}
In this part, we extract keyframes based on the LiDAR-Inertial odometry and extract all SRP from the LiDAR scan in the keyframe coordinate system, find the correspondence in the entire graph and construct constraints as demonstrated in Fig. \ref{factor_graph}.
\subsection{SRP Extraction}
For the calculation efficiency, we select keyframes as vertices of the factor graph according to the odometry of the front-end. Since we are using a  LiDAR based on scanning mechanism, the change of the yaw angle does not affect the selection of keyframes. The new keyframe will be selected only when the distance between the new frame and the previous keyframe exceeds 1m or the pitch angle or roll angle exceeds 10$\degree$.

We extract all SRP from the corrected LiDAR scan $ \mathcal{S}_i $ for each newly added keyframe $ K_i $. Here we define the plane as $ \pi(\mathbf{n},d) $ through the Hesse normal form described by Eq. (\ref{hesse}). $\mathbf{n} = [n_x, n_y, n_z]^T$ represents the unit normal vector of the plane, and $d$ represents the distance from the coordinate origin of $K_i$ to the plane. Next, apply RANSAC\cite{fischler1981random} to extract planes for $\mathcal{S}_i$, but not all planes are reserved for building constraints, but only those planes that can represent the structure of the building (e.g., ground, walls, etc.) are selected. Here we adopt the following strategies for the extraction of SRP:

\begin{itemize}
\item{Keep all the planes with more than $\mathcal{N}$ points (Here, we set $\mathcal{N}$ to $400$).}
\item{According to the normal vector of the extracted plane, three planes containing the most points and almost orthogonal are retained.}
\item{Use $80\%$ of the points in $\mathcal{S}_i$ to extract the plane, and the remaining points default to the unextractable points.}
\end{itemize}

Too many planes are extracted will increase the uncertainty of the RANSAC process and cause mismatches in the plane matching process. Here we only use three orthogonal planes to obtain the precise pose of the LiDAR with 6-DOF. At the same time, fewer edges will be constructed in the factor graph to reduce the calculation time.

\subsection{SRP Global Constraint}
To construct the global constraint, all SRP extracted from keyframe $K_i$ will be checked whether they have appeared in the previous keyframes. Here we denote all the planes added to the graph as $\boldsymbol{\Pi}=\left\{{\pi}_{1}^{K_0}, \cdots,{\pi}_{k_0}^{K_0} ,\cdots,{\pi}_{1}^{K_{i-1}}, \cdots, {\pi}_{k_{i-1}}^{K_{i-1}}\right\}$, and the SRP under the $K_i$ frame as $\boldsymbol{\Pi}^{K_i}=\left\{{\pi}_{1}^{K_i},\cdots,{\pi}_{k_i}^{K_i}\right\}$. First, according to the optimized results $\mathbf{T}^W_{K_m},m\in\{1,\cdots,i-1\}$ and the front-end odometry $\mathbf{T}^{K_i}_{K_{i-1}} $, the planes in $\boldsymbol{\Pi}$ are transformed to the frame of keyframe $ K_i $.

\begin{align} \label{transform-plane}
    \begin{split}
    \mathbf{T}^{K_i}_{K_m} &= \mathbf{T}^W_{K_m} \mathbf{T}^{K_{i-1}}_W \mathbf{T}^{K_i}_{K_{i-1}} = \begin{bmatrix} \mathbf{R}_{K_m}^{K_i} & \mathbf{p}_{K_m}^{K_i} \\ 
            \mathbf{0} &  1 \end{bmatrix}\\
        \begin{bmatrix} ^{\prime}{\mathbf{n}}^{K_i} \\ 
             ^{\prime}{d}^{K_i} \end{bmatrix} &= \begin{bmatrix} \mathbf{R}_{K_m}^{K_i} & 0 \\ 
            -{\mathbf{p}_{K_m}^{K_i}}^{T} &  1 \end{bmatrix}
            \begin{bmatrix} {\mathbf{n}}^{K_m} \\ 
             {d}^{K_m} \end{bmatrix}
    \end{split}
\end{align}

For all ${\pi}^{K_i}_{m}(\mathbf{n}^{K_i}, d^{K_i})\in\boldsymbol{\Pi}^{K_i},m\in\left\{1,\cdots,k_i\right\}$, calculate the angle $\delta\theta$ between its normal vector $\mathbf{n}^{K_i}$ and $^{\prime}{\mathbf{n}}^{K_i}$ and the distance $\delta d$ between $d^{K_i}$ and $^{\prime}{d}^{K_i}$. Once $\delta \theta$ and $\delta d$ are lower than the preset threshold, add a plane edge to the factor graph. Otherwise, it's considered a new plane and added to $\boldsymbol{\Pi}$.

\subsection{Graph optimization}
When the LiDAR-Inertial odometry and SRP construct the constraints between keyframes, the SLAM problem is expressed in a factor graph. The vertices of the graph represent states of being optimized, and the edges represent the constraints formed by the sensors' measurements, as shown in Fig. \ref{factor_graph}. Following\cite{kummerle2011g, grisetti2010tutorial}, the maximum likelihood estimation problem can be expressed as this nonlinear least-squares problem:
\begin{align}
    \begin{split}\label{grapg-opti-func}
        F(\mathbf{x})=\sum_{\langle i, j\rangle \in \mathcal{C}} \mathbf{e}\left(\boldsymbol{x}_{i}, \boldsymbol{x}_{j}, \boldsymbol{z}_{i j}\right)^{T} \boldsymbol{\Omega}_{i j} \mathbf{e}\left(\boldsymbol{x}_{i}, \boldsymbol{x}_{j}, \boldsymbol{z}_{i j}\right)
    \end{split}
\end{align}
where $\mathbf{x}$ represents all states to be optimized and $\boldsymbol{x}_i, \boldsymbol{x}_j \in \mathbf{x}$, $\boldsymbol{z}_{ij}$ and $\boldsymbol{\Omega}_{ij}$ represent the mean and the information matrix of a constraint between $\boldsymbol{x}_i$ and $\boldsymbol{x}_j$, $\mathcal{C}$ is the set of pairs of indices for which the constraint exist, and $\mathbf{e}\left(\boldsymbol{x}_{i}, \boldsymbol{x}_{j}, \boldsymbol{z}_{i j}\right)$ is the error function between $\boldsymbol{x}_{i}$, $\boldsymbol{x}_{j}$ and $\boldsymbol{z}_{i j}$. Eq. (\ref{grapg-opti-func}) can be minimized by Gauss-Newton or Levenberg-Marquardt algorithm.

\section{EXPERIMENTS}

\begin{figure}[!t]
\centering
\includegraphics[width=3in]{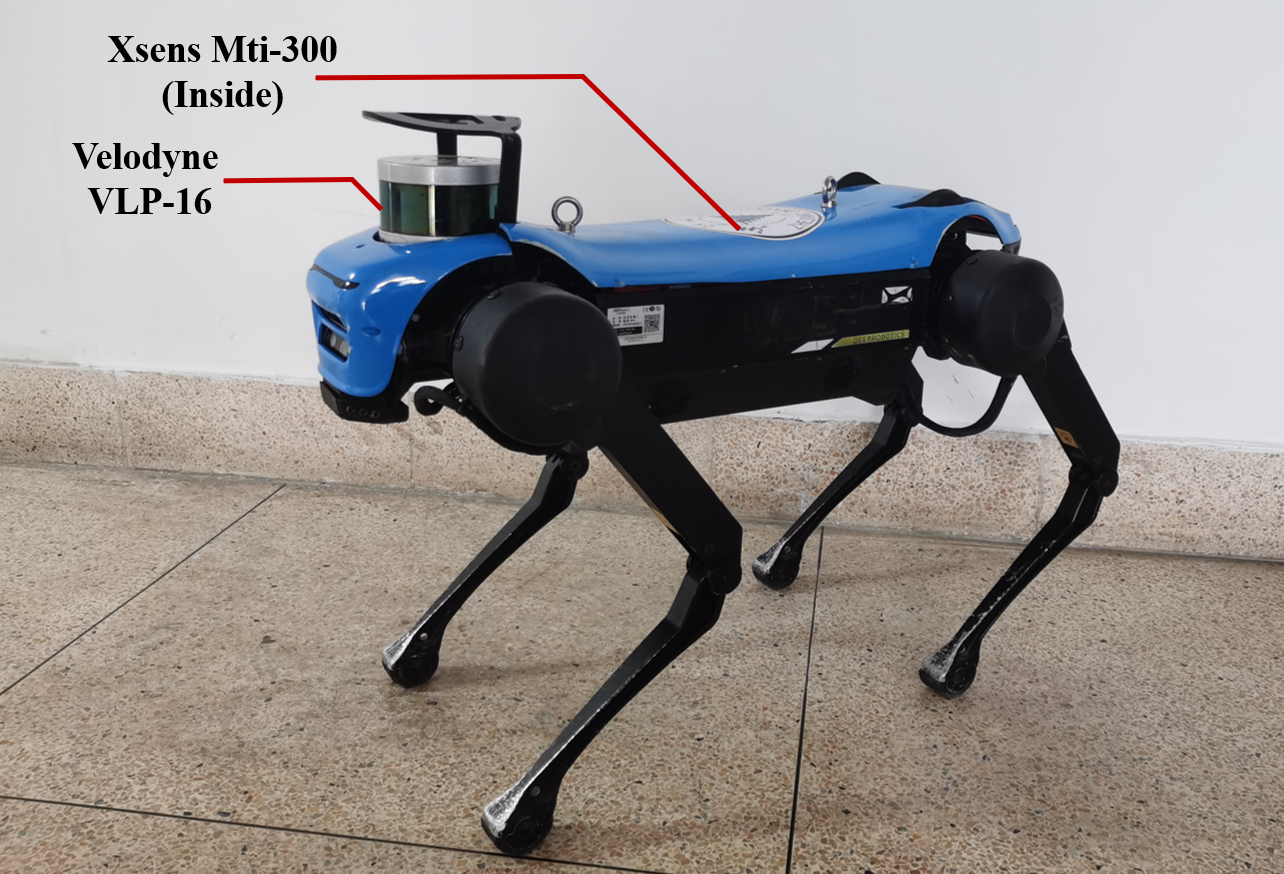}
\caption{The Jueying Mini quadruped robot.}
\label{jueying}
\end{figure}

\subsection{Experimental Settings}

To verify the versatility of the algorithm, we conducted experiments in different buildings. We used the Jueying Mini robot (Fig. \ref{jueying}) equipped with Velodyne VLP-16 and Xsens Mti-300 to collect data from multiple sets of multi-story scenes. The LiDAR is fixed on the head of Jueying, and the IMU is assembled at the center of mass. The LiDAR operates at a frequency of 10 Hz, and the IMU outputs orientation, angular velocity, and linear acceleration at 400 Hz. Since there are currently no publicly available datasets of LiDAR and IMU for indoor multi-story scenes, we used Jueying to collect actual data in two buildings and named them \textit{Building A} and \textit{Building B}, respectively. \textit{Building A} is a five-story building in the shape of long corridor, and \textit{Building B} is a six-story building with two long corridor-shaped scattered on the left and right. Our algorithm is tested on a NUC mini PC with Intel Core i7-7567U, 16G memory.

\subsection{Results and Analysis}
In this section, we provide the results of our experiments. We compared the state-of-the-art SLAM algorithms based on multi-sensor fusion, including LIO-Mapping\cite{ye2019tightly}, Fast-LIO2\cite{xu2021fast}, and LIO-SAM\cite{shan2020lio}, and provided the mean running time of each part of our algorithm. These three are perfect LiDAR-Inertial SLAM algorithms that can apply to most scenarios. Still, in particular scenarios such as indoor multi-story, they may not achieve the best results. Due to the unique experimental scene, we cannot obtain the ground truth of the robot motion. At the same time, we set the robot's starting point and ending point to be the same when collecting data to calculate the relative position and orientation deviation. In the case of not adding loop closure, we can use this criterion to judge the accuracy of all algorithms.


\begin{figure*}[!t]
\centering
\subfigure[]{\includegraphics[width=3.5in]{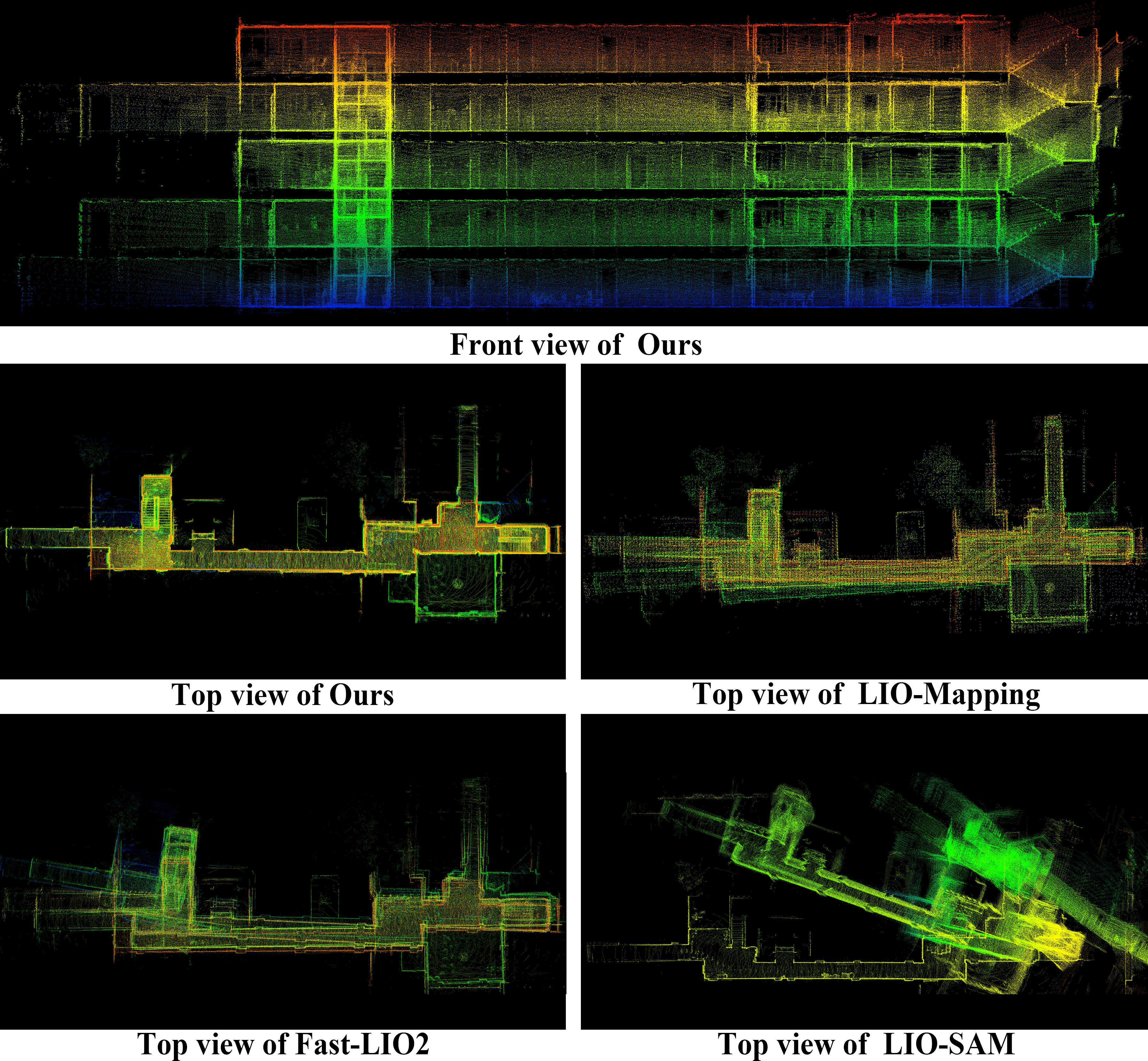}}
\subfigure[]{\includegraphics[width=3.5in]{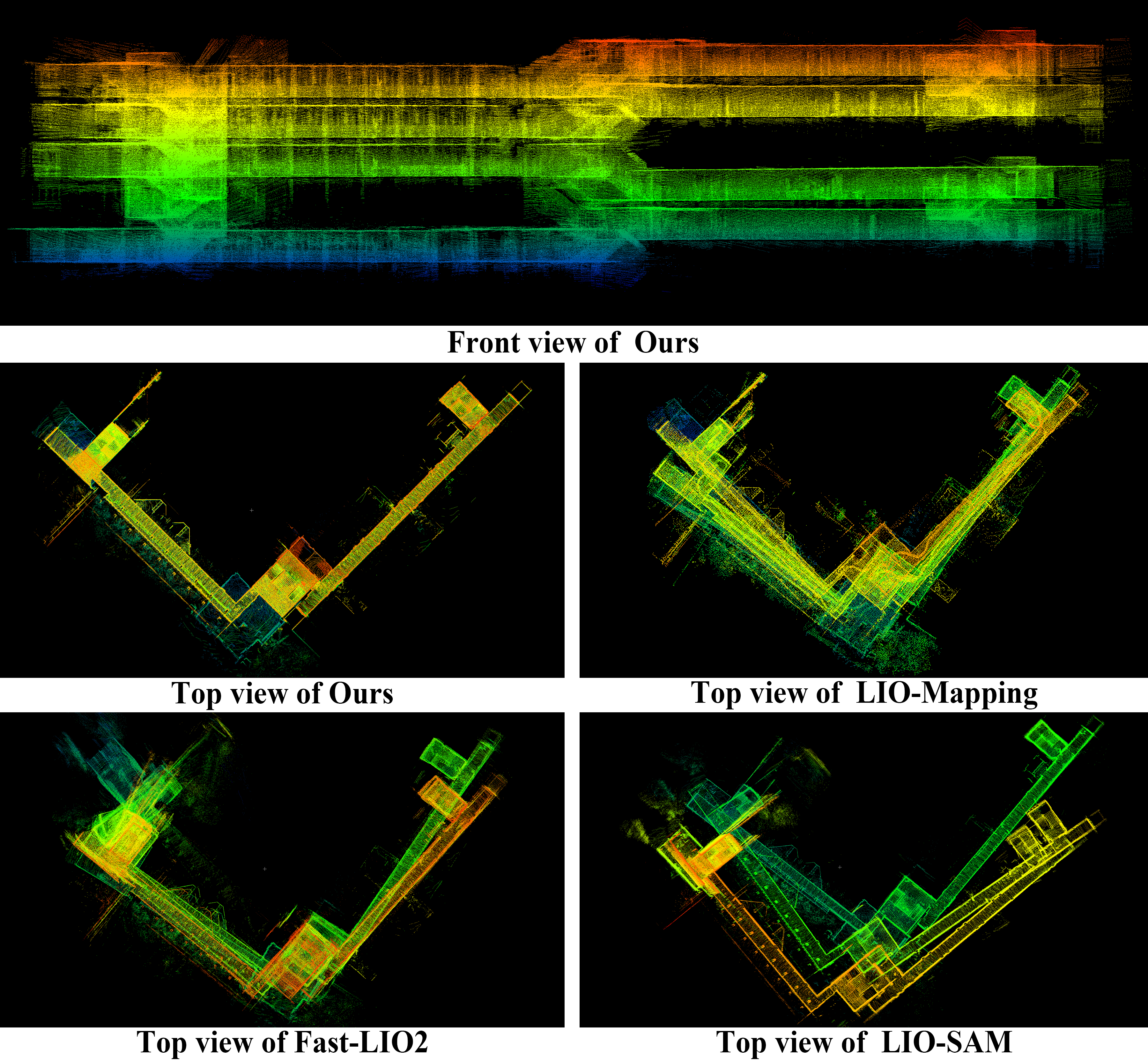}}
\caption{Maps generated by Ours, LIO-Mapping, FAST-LIO2, and LIO-SAM. The other three algorithms drift on different stories, except ours. (a) Maps of Building A. (b) Maps of Building B.}
\label{pcd}
\end{figure*}

\begin{figure*}[!t]
\centering
\subfigure[]{\includegraphics[width=3.5in]{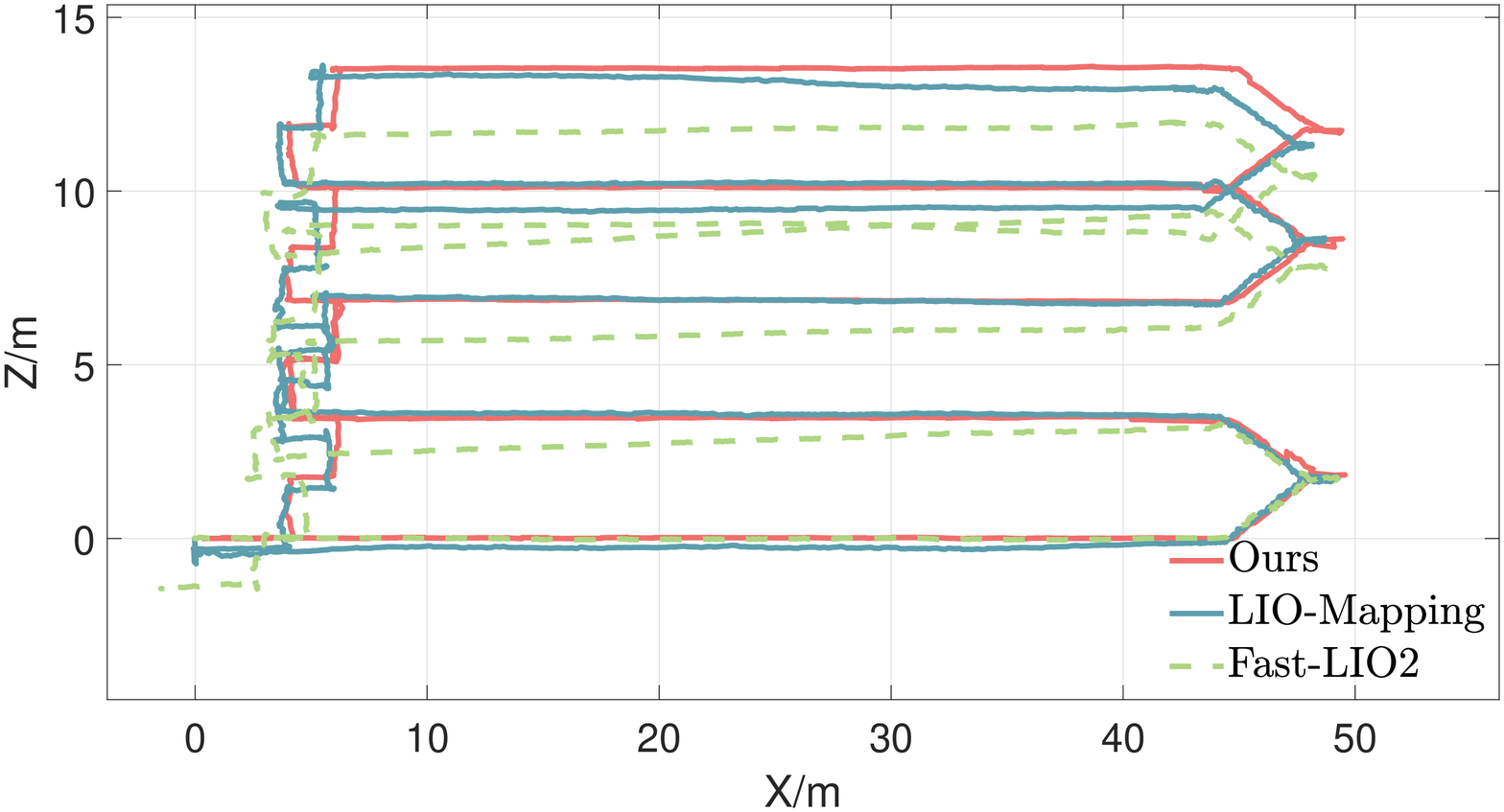}}
\subfigure[]{\includegraphics[width=3.5in]{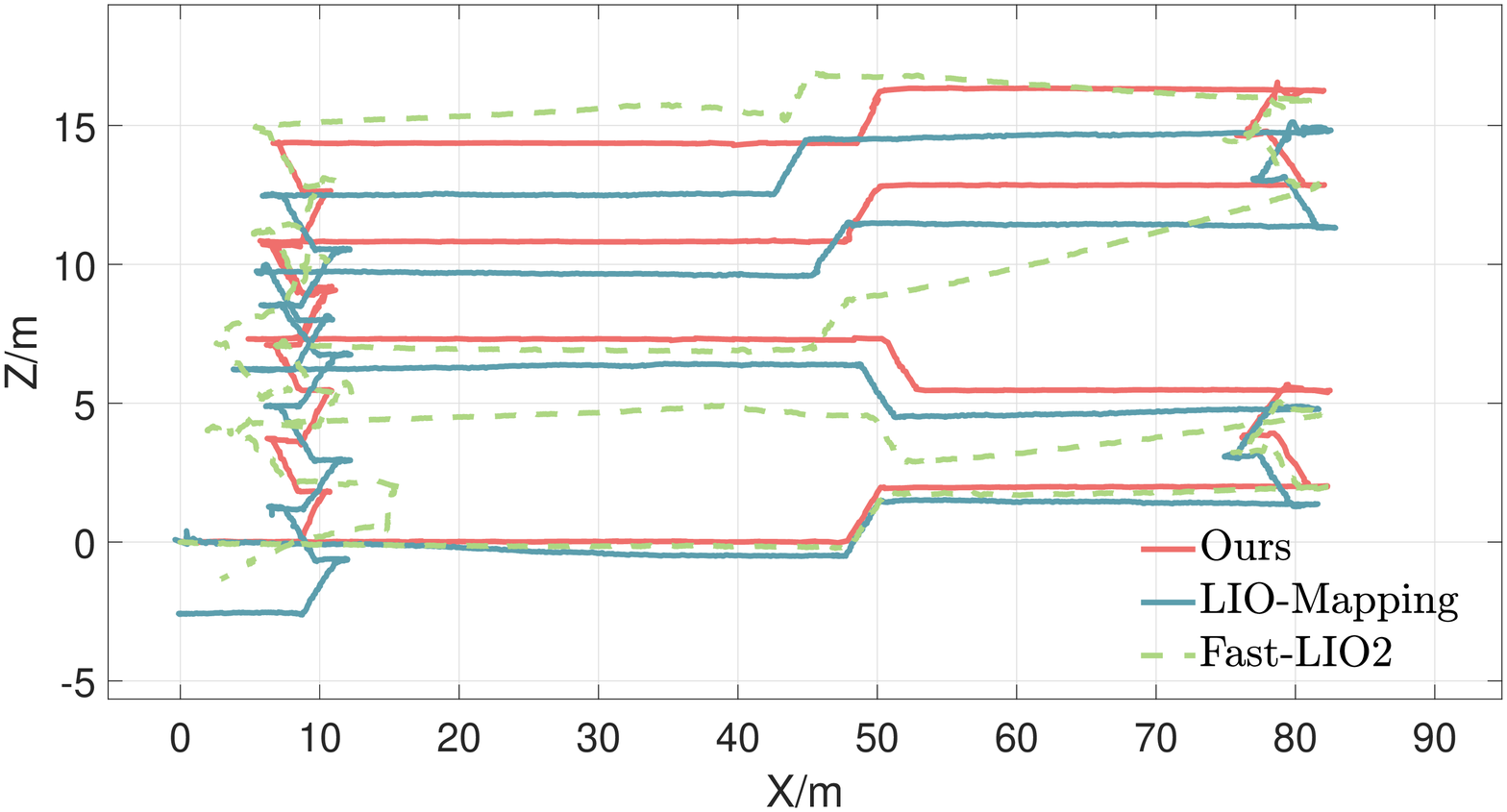}}
\\
\centering
\subfigure[]{\includegraphics[width=3.5in]{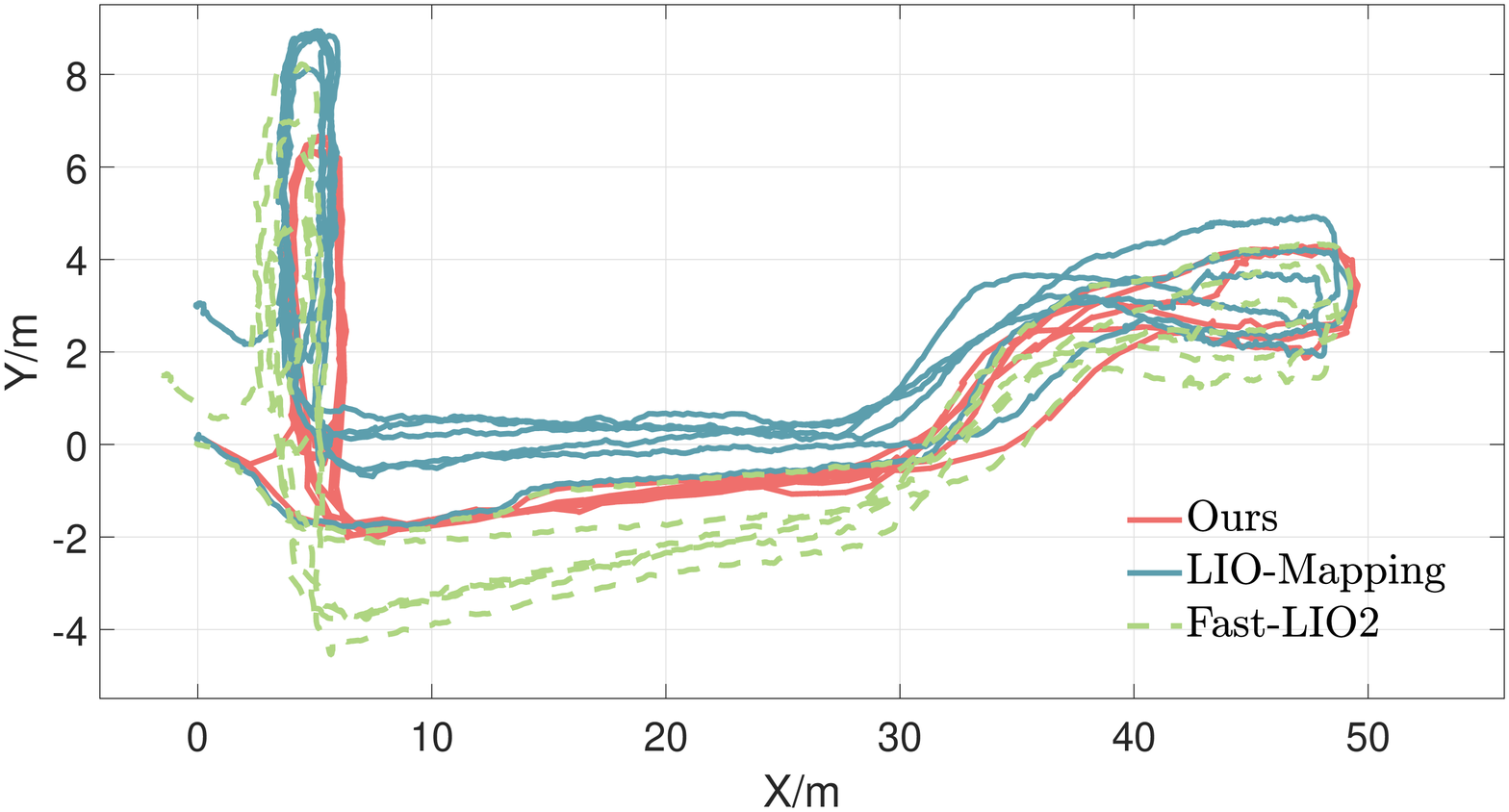}}
\subfigure[]{\includegraphics[width=3.5in]{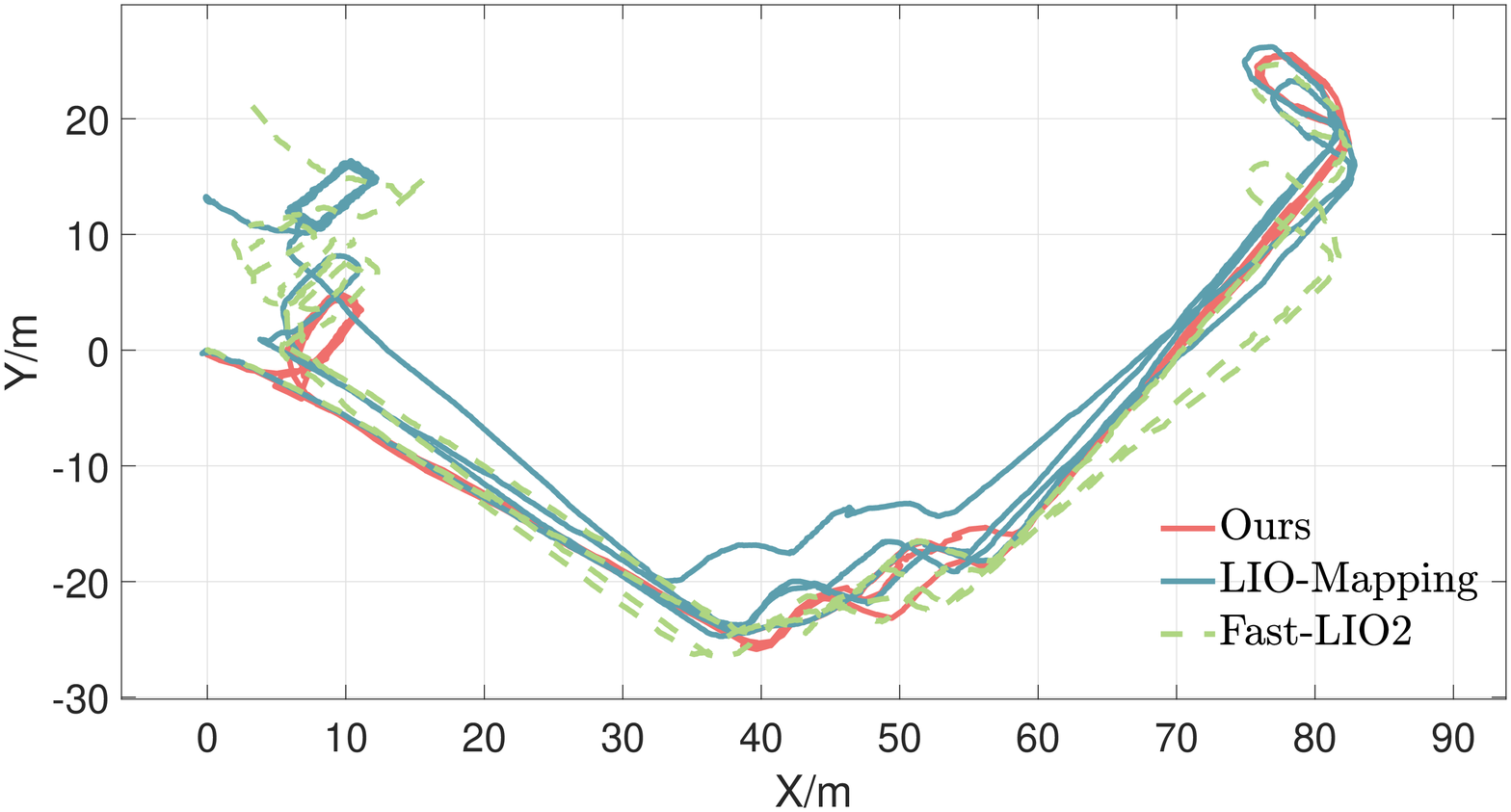}}
\caption{Comparison of trajectories estimated by different algorithms (LIO-SAM fails on both datasets, so we don't plot its trajectory.). (a) Front view of trajectories in Building A. (b) Front view of trajectories in Building B. (c) Top View of trajectories in Building A. (d) Top View of trajectories in Building B.}
\label{traj plot}
\end{figure*}

\textbf{Overview} The performance of four algorithms on the \textit{Building A} and \textit{Building B} datasets are shown in Fig. \ref{pcd}. We can see that our algorithm is significantly better than the other three algorithms on both datasets because of plane constraints. When the 16-line LiDAR moves horizontally, the number of points scanned on the ground is relatively minor. Therefore, during the scan-to-scan registration, the height estimation will produce more significant deviations, especially in degraded scenarios such as corridors. Despite the aid of IMU, there will still be cumulative errors, which can be seen more clearly in FAST-LIO2 and LIO-SAM. LIO-Mapping optimizes each state in the sliding window, which consumes more time, so the effect of height estimation is better, but in the end, it does not return to the starting point as well. The other three algorithms did not return to the starting point in the end due to the lack of performing loop closure.

\textbf{Trajectory} Fig. \ref{traj plot} shows the trajectories of three algorithms in two datasets. LIO-SAM failed in \textit{Building A} and \textit{Building B}, so we did not plot its trajectory. Although we do not have global ground truth, we can see in Fig. \ref{traj plot}(b) and Fig. \ref{traj plot}(d) that in the staircase on the left of \textit{Building A} and \textit{Building B}, the other two algorithms drift a lot. Still, after adding plane constraints, ours can maintain the consistency of different floors. Table \ref{tab:error} provides the relative deviations of translation and rotation. Since accurate state estimation is achieved on other stories, our algorithm can return to the starting point without loop closure. Because of the same planes used to construct constraints, both translation and rotation are almost consistent with the starting point. Table \ref{tab:time} lists the running times of different parts of our algorithm. Since an accurate front-end odometry can prevent mismatching in the plane matching process, it takes a little longer to estimate each state to ensure accuracy.

\begin{table*}[!t]
\renewcommand\arraystretch{1.2}
\caption{Relative deviation of different SLAM algorithms under the same starting point and ending point.}
    \centering
    \begin{tabular}{c | c |c |c c c c |c c c c}
    \toprule
    \multirow{2}{*}{\textbf{Dataset}}& \multirow{2}{*}{\textbf{Distance}} & \multirow{2}{*}{\textbf{System}}  & \multicolumn{4}{c|}{\textbf{Translation (m)}} & \multicolumn{4}{c}{\textbf{Rotation (rad)}}\\
         & \textbf{(m)}& & \textbf{$\Delta $X} &\textbf{$\Delta $Y} & \textbf{$\Delta $Z} & \textbf{$\Delta $XYZ} & \textbf{$\Delta $Yaw} & \textbf{$\Delta $Pitch} & \textbf{$\Delta $Roll} & \textbf{$\Delta $Angle}\\
    \hline
        \multirow{4}{*}{{Building A}} & \multirow{4}{*}{{396}} & Ours &  \textbf{0.018} & \textbf{0.023} & \textbf{0.015} & \textbf{0.033} & \textbf{-0.021} & \textbf{-0.002} & \textbf{0.018} & \textbf{0.028} \\
        & & LIO-Mapping  & -0.073 & 3.001 & -0.305 & 3.017 & 0.114 & -0.019 & -0.018 & 0.117 \\
        & & FAST-LIO2 & -1.529 & 1.424 & -1.442 & 2.539 & -0.078 & -0.023 & -0.086 & 0.118 \\
        & & LIO-SAM & -8.576 & -35.110 & -25.802 & 44.407 & 1.629 & -0.629 & -1.438 & 2.262 \\
    \hline
        \multirow{4}{*}{{Building B}} & \multirow{4}{*}{{613}} & Ours  & \textbf{0.021}
 & \textbf{0.023} & \textbf{0.002} & \textbf{0.031} & \textbf{-0.006} & \textbf{-0.007} & \textbf{-0.006} & \textbf{0.011} \\
        & & LIO-Mapping  & 0.412 & 12.857 & -2.588 & 13.121 & -0.036 & -0.008 & -0.038 & 0.054 \\
        & & FAST-LIO2  & 2.943 & 21.479 & -1.328 & 21.720 & -0.356 & -0.176 & -0.096 & 0.409 \\
        & & LIO-SAM & 4.296 & 12.792 & 12.109 & 18.131 & 2.409 & 0.030 & -0.180 & 2.416 \\
    \bottomrule
    \end{tabular}
    \label{tab:error}
\end{table*}

\begin{table*}[!t]
\renewcommand\arraystretch{1.2}
\caption{Mean running time (ms) of different components.}
    \centering
    \begin{tabular}{c| c| c c| c}
    \toprule
        \multirow{2}{*}{\textbf{Dataset}} & \multirow{2}{*}{\textbf{LiDAR-Inertial Odometry}} & \multicolumn{2}{c|}{\textbf{SRP Constraint}} & \multirow{2}{*}{\textbf{Back-End Optimization}}\\
        &  & \textbf{SRP Extraction} & \textbf{Find Global Constraint} \\
    \hline
        Building A & 249.10 & 413.34 & 4.90 & 252.80 \\
        Building B & 316.14 & 301.08 & 8.41 & 353.53 \\
    \bottomrule
    \end{tabular}
    \label{tab:time}
\end{table*}

\section{CONCLUSION and FUTURE WORK}

This paper proposes a SLAM algorithm for indoor multi-story scenes with a plane as the main feature. We use the tightly coupled LiDAR and IMU as the front-end and optimize the two simultaneously by constructing the error function of IMU pre-integration and scan registration to obtain more accurate odometry information. We search for SRP in keyframes instead of all planes at the back-end, which makes it construct fewer but more significant edges in the factor graph. According to the normal direction of the plane and the distance to the origin, it searches and matches with global SRP and constructs constraints. This allows the robot to build the same constraints on different floors, eliminates the cumulative error, and achieves an effect similar to "dimensionality reduction." Experiments show that our algorithm can significantly improve the state estimation. The constructed global map records the structural characteristics of the building well and is better than the state-of-the-art SLAM algorithm, LIO-Mapping\cite{ye2019tightly}, Fast-LIO2\cite{xu2021fast}, and LIO-SAM\cite{shan2020lio} in the indoor multi-story scenario. However, the current plane matching process relies heavily on front-end odometry. If the front-end odometry drifts a lot, the plane could be mismatched. Therefore, our algorithm takes a lot of time in the previous stage, which causes it to not run in real-time. In the future, it may be necessary to combine features unique to the plane to make the matching process more robust.


\bibliographystyle{./IEEEtran}
\bibliography{./IEEEabrv, ./mybibfile}

\newpage
 
\vspace{11pt}

\vfill

\end{document}